\pdfoutput=1

\documentclass[11pt]{article}

\usepackage{acl}

\usepackage{times}
\usepackage{latexsym}

\usepackage[T1]{fontenc}

\usepackage[utf8]{inputenc}

\usepackage{microtype}

\usepackage{inconsolata}

\usepackage{graphicx}
\usepackage{float}

\usepackage{booktabs}
\usepackage{multirow}
\usepackage{array}      
\usepackage{adjustbox}  
\usepackage{makecell}
\usepackage[normalem]{ulem}
\useunder{\uline}{\ul}{}

\usepackage{amsmath}
\usepackage{algorithm}
\usepackage{algpseudocode}
\usepackage{amsfonts}
\usepackage{bm}

\title{Optimal Brain Iterative Merging: Mitigating Interference in LLM Merging}

\author{Zhixiang Wang $^{1,2,3,*}$ \quad
Zhenyu Mao $^{2,}$\thanks{\ \ \ Equal contribution.} \quad
Yixuan Qiao $^{2}$ \quad
Yunfang Wu $^{1,\dag}$ \quad
Biye Li $^{2,}$\thanks{\ \ \ Corresponding author.} \\
  $^{1}$National Key Laboratory for Multimedia Information Processing, Peking University \\
  $^{2}$GAI, Du Xiaoman \quad
  $^{3}$School of Software \& Microelectronics, Peking University \\
  \texttt{ekko@stu.pku.edu.cn} \quad
  \texttt{mao\_zy@126.com} \\
  \texttt{wuyf@pku.edu.cn} \quad
  \texttt{\{qiaoyixuan,xiangyi\}@duxiaoman.com}
}

\begin{document}
\maketitle
\begin{abstract}
Large Language Models (LLMs) have demonstrated impressive capabilities, but their high computational costs pose challenges for customization. Model merging offers a cost-effective alternative, yet existing methods suffer from interference among parameters, leading to performance degradation. In this work, we propose \textbf{O}ptimal \textbf{B}rain \textbf{I}terative \textbf{M}erging (OBIM), a novel method designed to mitigate both intra-model and inter-model interference.
OBIM consists of two key components: (1) A \textit{saliency measurement mechanism} that evaluates parameter importance based on loss changes induced by individual weight alterations, reducing intra-model interference by preserving only high-saliency parameters. (2) A \textit{mutually exclusive iterative merging framework}, which incrementally integrates models using a binary mask to avoid direct parameter averaging, thereby mitigating inter-model interference.
We validate OBIM through experiments on both Supervised Fine-Tuned (SFT) models and post-pretrained checkpoints. The results show that OBIM significantly outperforms existing merging techniques. 
Overall, OBIM provides an effective and practical solution for enhancing LLM merging. We will publicly release our code upon the acceptance of this paper.
\end{abstract}

\section{Introduction}


Existing research \cite{evolutionary, controlled, wan2024knowledge} has demonstrated that a composite LLM can be constructed by merging the parameters of different expert LLMs. Traditional approaches \cite{modelsoups, matena2022merging, jin2022dataless} employ matrices to determine task-specific coefficients and perform a weighted average based on these coefficients. Methods grounded in task arithmetic \cite{ta} leverage task vectors, defined as the difference between the parameter values of a fine-tuned model and those of its pre-trained counterpart, to effectively manipulate and integrate the knowledge embedded within the models.

State-of-the-art model merging methods \cite{ties, tallmask, dare} have shown that task performance degradation is primarily caused by interference between parameter values, as aggregation operations, such as averaging, can alter the parameter distribution \cite{yu2024extend}. The interference can be categorized into two types: intra-model interference and inter-model interference.

\begin{figure}[t]
\centering 
\includegraphics[width=0.9\columnwidth]{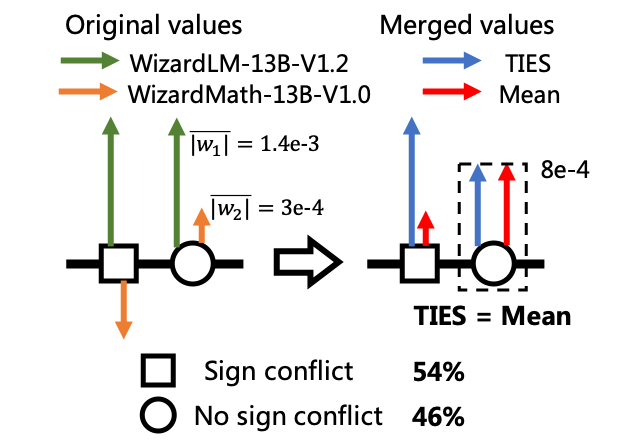}
\caption{Illustration of inter-model interference. The dotted box highlights cases where TIES fails to resolve interference. Approximately $46\%$ of parameters deviate from the original models due to task vector averaging in the absence of sign conflicts.}
\label{fig:intro} 
\end{figure}

Intra-model interference arises from redundant parameters within a single model. Due to the over-parameterized nature of neural networks \cite{choudhary2020comprehensive, he2023structured}, removing a significant portion of the parameters often has little impact on model performance \cite{sun2023simple, kim2024shortened}. However, these redundant parameters introduce noise during the model merging process, adversely affecting the outcome. To address this, it is crucial to identify parameters that are closely related to the target task. Existing approaches, however, primarily rely on magnitude-based methods, assuming that parameter magnitude directly correlates with saliency. For instance, TIES \cite{ties} trims the parameters with the smallest magnitudes, the Model Breadcrumbs \cite{model-breadcrumbs} highlight the importance of removing the parameters with the largest weights to further reduce noise.
While these methods demonstrate effectiveness, they fall short of fully revealing the true saliency of the parameters.

Interference between models arises due to variations in parameter distributions \cite{slerp, jang2024model}. Directly averaging these parameters can lead to performance degradation. TIES addresses this issue by resolving sign conflicts in parameter values, aligning them based on the direction of the largest total movement across models. Similarly, TALL-Mask \cite{tallmask} is designed to exclude parameters that are relevant only to a subset of tasks. While these methods effectively mitigate inter-model interference under certain conditions, their effectiveness diminishes when parameter distributions deviate from expected patterns, causing them to revert to simple averaging. As shown in Figure~\ref{fig:intro}, when there is no sign conflict, TIES yields the same result as simple averaging, deviating from both input models and leading to suboptimal performance.

To address the interference problem in model merging, we propose a novel method for LLMs called Optimal Brain Iterative Merging (OBIM). Our approach comprises two core components: a saliency measurement mechanism to filter intra-model interference and a mutually exclusive iterative merging framework to prevent inter-model interference.

In detail, our approach measures the saliency of parameters within a single model by evaluating the loss change induced by altering each parameter. Inspired by layer-wise model pruning methods \cite{obc, gptq}, we forgo reliance on the overall model loss during training and instead independently apply the Mean Square Error (MSE) to each linear weight. This enables calculation of the output distribution difference between the trained weight and the original weight, providing a more precise and efficient measure of parameter saliency. By retaining parameters with high saliency, we effectively reduce intra-model interference in model merging.

Subsequently, we design an iterative merging framework to integrate models step by step in a mutually exclusive manner, mitigating inter-model interference. Specifically, we employ a binary mask to track the positions that have already been merged. At each step, parameters with the highest saliency, which are not yet recorded in the mask, are selected from a model and integrated into the base model. This ensures that each position is occupied by only one parameter, thereby eliminating the need for averaging operations.



In summary, we propose a novel method, OBIM, to mitigate both intra-model and inter-model interference in LLM merging. To validate the effectiveness of our method, we conducted model merging experiments on Supervised Fine-Tuned (SFT) models of Llama2 \cite{llama2} for multi-task merging and post-pretrained checkpoints of Qwen2 \cite{qwen2} for catastrophic forgetting recovery. The results of both experiments demonstrate that OBIM significantly outperforms existing approaches. In addition, extensive ablation studies and analyses of key factors provide a comprehensive understanding of OBIM.

\begin{figure*}[ht]
\centering 
\includegraphics[width=0.9\textwidth]{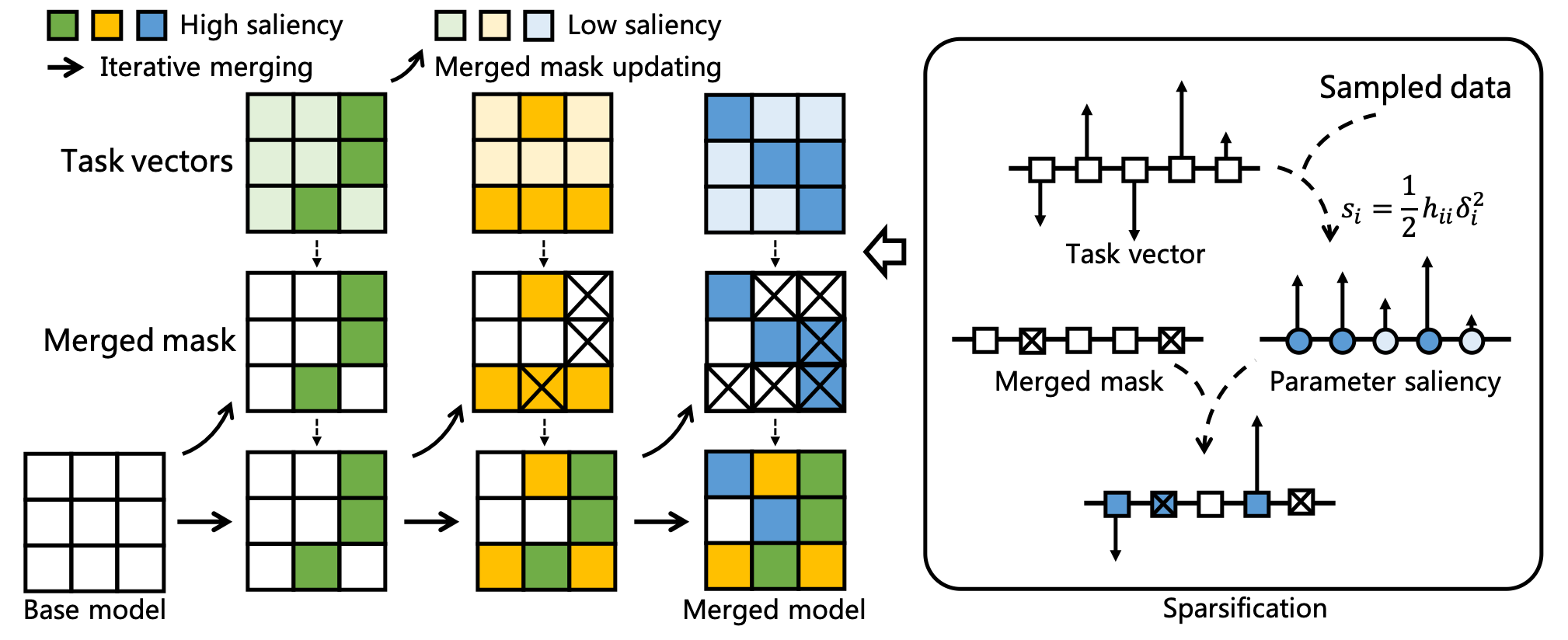}
\caption{An overview of the proposed method. The left part depicts the iterative merging process, while the right part details how parameters are selected at each iteration step through the cooperation of parameter saliency and the merged mask.
}
\label{Fig.main} 
\end{figure*}

\section{Preliminaries}

\subsection{Model Merging Problem}
In this paper, we focus on merging models that are optimized from the same backbone.  
Given $K$ models with parameters $\{\theta^{1}, \theta^{2}, \dots, \theta^{K}\} \in \mathbb{R}^d$, each trained on a distinct task or setting $\{t_1, t_2, \dots, t_K\}$ from a shared base model $\theta^{B} \in \mathbb{R}^d$. Model merging aims to fuse these parameters into a single model with parameters $\theta^M \in \mathbb{R}^d$, and enable $\theta^M$ to effectively handle all $K$ tasks simultaneously.

\subsection{Task Vector}  
A task vector $\delta^k \in \mathbb{R}^d$ for model $\theta^{k}$ is defined as the delta weights between the trained model's parameters and those of the backbone:  
\begin{equation}  
\delta^k = \theta^k - \theta^B,  
\end{equation}  
which represents both the direction and magnitude of parameter updates during training. 

By merging the task vectors $\{\delta^1, \delta^2, \dots, \delta^K\}$ of $K$ models into a single task vector $\delta^M$, the parameters of the merged model can be expressed as  
\begin{equation}  
\theta^M = \theta^B + \delta^M.  
\end{equation}

\section{Methodology}

The proposed merging method comprises two key components: Optimal Brain Merging (OBM), a saliency-based mechanism that selects high-saliency parameters for merging, and Iterative Merging (IM), an iterative framework designed to mitigate interference between models. Figure~\ref{Fig.main} illustrates the complete merging process of our method.

\subsection{Optimal Brain Merging}

Optimal Brain Damage (OBD) \cite{obd} and its subsequent works \cite{obs, obc} aim to establish effective criteria for pruning or quantizing specific weights while minimizing the impact on model performance. The fundamental idea is to leverage the second derivative of the objective function with respect to the parameters to compute their "saliencies." Building upon the idea, we introduce \textbf{O}ptimal \textbf{B}rain \textbf{M}erging (OBM) to mitigate intra-model interference by identifying and eliminating negligible delta weights in task vectors.

Given a trained LLM with parameters $\theta$ and task vector $\delta$, our goal is to identify a subset of parameters in the task vector whose removal results in minimal increase in the objective function $\mathcal{L}$. The change in the objective function is measured using a Taylor series expansion:  
\begin{equation}
\begin{aligned}
    \Delta \mathcal{L} = & \sum_i\frac{\partial\mathcal{L}}{\partial\theta_i}\delta_i + \frac{1}{2}\sum_{i}\frac{\partial^2\mathcal{L}}{\partial\theta_i^2}\delta_i^2 \\ 
    & + \frac{1}{2}\sum_{i \neq j}\frac{\partial^2\mathcal{L}}{\partial\theta_i\theta_j}\delta_i\delta_j + O(||\delta||^3).
\end{aligned}
\end{equation}
Assuming that $\mathcal{L}$ is at a local minimum and that each parameter contributes to $\Delta \mathcal{L}$ independently, the first derivative, the off-diagonal terms of the second derivative, and the higher-order terms can be discarded. Consequently, $\Delta \mathcal{L}$ can be approximated as:  
\begin{equation}
    \Delta \mathcal{L} \approx \frac{1}{2}\sum_{i}\frac{\partial^2\mathcal{L}}{\partial\theta_i^2}\delta_i^2.
\end{equation}

The change in $\Delta \mathcal{L}$ when removing the parameter at position $i$ indicates how much it affects performance, thereby representing its saliency:  
\begin{equation}
    s_i = \frac{1}{2} \frac{\partial^2\mathcal{L}}{\partial\theta_i^2} \delta_i^2 = \frac{1}{2} h_{ii} \delta_i^2,
\end{equation}
where $h_{ii}$ denotes the $i$-th diagonal element of the Hessian matrix of the loss for the given model.  
Parameters with low saliency, which contribute minimally to $\Delta \mathcal{L}$, should be removed. 

However, computing the Hessian matrix requires a back-propagation process through the LLM if the objective function used during LLM training is considered \cite{bowen2024beyond}. To avoid the high computational cost comparable to model training, we take inspiration from layer-wise pruning approaches \cite{obc, gptq} and employ the Mean Squared Error (MSE) as the objective function for each linear layer independently. 

Formally, let $\mathbf{X}^l$ be the input to the $l$-th layer with the weight matrix $\mathbf{W}^l$. The objective is defined as:  
\begin{equation}
    \Delta \mathcal{L}^l = \left\| \mathbf{W}^l \mathbf{X}^l - \mathbf{W}^{B} \mathbf{X}^l \right\|_2^2 = \left\| \Delta \mathbf{W}^l \mathbf{X}^l \right\|_2^2,
\end{equation}
where $\mathbf{W}^{B}$ is the corresponding layer weight of the base model, and $\Delta \mathbf{W}^l$ is the task vector of the layer. To approximate $\mathbf{X}^l$, we take the mean over a small set of input samples. This function measures the squared distance between the output of the trained weights and the original weights.

The Hessian matrix under the layer-wise MSE loss is computed as $\mathbf{H}^l = 2\mathbf{X}^l {\mathbf{X}^l}^{\top}$. Thus, we only need to perform forward propagation of the LLM to obtain the input for each layer, enabling the computation of parameter saliencies. Moreover, forward propagation does not require any labels or targets, only the input portion of the samples is needed. Beyond its simplicity, layer-wise saliency provides a more precise and accurate measure of parameter importance within each layer. For non-linear layers, such as bias layers, we apply random pruning for parameter sparsification.

\subsection{Iterative Merging}

To address inter-model interference, we propose a merging framework called \textbf{I}terative \textbf{M}erging (IM). This method iteratively updates a unique, non-overlapping subset of parameters from each task vector, preventing weight interference among task vectors.

For each task vector $\delta^k$ to be merged, a binary mask $\mathcal{M}(P_k)$ is constructed to satisfy the following constraints:
\begin{equation}
\begin{aligned}
    \mathcal{M}(P_k)_i &= 
    \begin{cases}
        1, & \text{if } i \in P_k, \\
        0, & \text{otherwise}.
    \end{cases} \\
    \text{subject to} \quad & \bigcup_{k=1}^K P_k \subseteq \{1, 2, \dots, d\}, \\
    & P_k \cap P_j = \emptyset \quad \text{for } k \neq j.
\end{aligned}
\end{equation}
Here, $P_k$ represents the set of indices corresponding to the parameters selected for merging from $\delta^k$, and $d$ denotes the total number of parameters in each $\delta^k$. Using the binary masks, the merging process is then formulated as:
\begin{equation}
    \theta^M = \theta^B + \sum_{k=1}^K\delta^k\cdot\mathcal{M}(P_k).
\end{equation}
Although the formulation involves summation, no direct addition occurs between different task vectors, as the binary masks ensure that each parameter index is selected at most once.

While there are many ways to construct non-overlapping binary masks for all models, we introduce a simple and easily controllable method by iteratively updating a merged mask to track the positions that have already been merged. Specifically, at the beginning, the merged mask is an empty set. At each step, starting with a task vector $\delta^k$, we first exclude the indices that are already in the merged mask. Then, we sort the remaining parameter indices of $\delta^k$ based on their saliencies, selecting the top $n_k\%$\footnote{We ensure that $\sum_k n_k \leq 1$.} of the indices to form $P_k$. Finally, we update the merged mask with $P_k$. The procedure is described in Algorithm~\ref{alg.IM}.

\begin{algorithm}
\renewcommand{\algorithmicrequire}{\textbf{Input:}}
\renewcommand{\algorithmicensure}{\textbf{Output:}}
\algnewcommand{\Initialize}[1]{\State \textbf{Initialize:} #1}
\caption{Iterative Merging}
\label{alg.IM}
\begin{algorithmic}[1]
    \Require Base model parameters $\theta^B$, total parameter count $d$, task vectors $\delta^{1:K}$, merging ratios $n_{1:K}\%$, saliency score sets $S_{1:K}$
    \Ensure Merged model $\theta^M$
    \Initialize{merged mask $M \gets \emptyset$, merging order $O \gets [o_1, o_2, \dots, o_K]$}
    \For{$k$ in $O$}
        \State $S_k \gets \{ s_i \mid i \notin M, s_i \in S_k \}$
        \State Sort $S_k$ in descending order
        \State $\hat{S_k} \gets$ Select the top $n_k\%$ elements of $S_k$
        \State $P_k \gets \{ i \mid s_i \in \hat{S_k}, i \in \{1, \dots, d\} \}$
        \State Update merged mask: $M \gets M \cup P_k$
    \EndFor
    \State \Return $\theta^M \gets \theta^B + \sum_{k=1}^K\delta^k\cdot\mathcal{M}(P_k)$
\end{algorithmic}
\end{algorithm}

In practice, we apply iterative merging to each layer independently, utilizing the task vector of each layer rather than the entire model. This approach not only improves memory efficiency but also enables the integration with OBM by leveraging its layer-wise saliency scores.

An important factor that significantly affects the performance of the merged model is the iteration order of the merging process. Since earlier merged models occupy parameter positions, highly salient parameters from later models may have limited opportunities to be incorporated. To address this issue, we utilize a rotation operation that dynamically shifts the merging order across different layers, preventing a few models from dominating the process. 
Formally, for layer $l$, we maintain a list to record the merging order of models: $O^l = [o_1, o_2, \dots, o_K]$ The merging order for layer $l+1$ is then updated by a left rotation operation: $O^{l+1} = \mathcal{LR}(O^l, 1)$. We further conduct an experiment to discuss how the iteration order influences the performance in Section~\ref{sec:exp_order}.

\subsection{Optimal Brain Iterative Merging}

OBM and IM can be combined with other existing methods. Taking TIES as an example, when combining TIES with OBM, the magnitude-based parameter pruning is replaced by a saliency-based approach. Similarly, when using TIES together with IM, we utilize global magnitude as saliency scores to construct the merged mask. However, the combination of OBM and IM, referred to as OBIM, yields better results. We conduct an ablation study to demonstrate the effectiveness of each component in Section~\ref{sec:ablationstudy}.

\section{Experiments}

\begin{table*}[htbp]
\centering
\resizebox{0.9\textwidth}{!}{%
\begin{tabular}{l|c|cc|cc|cc|c}
\hline
\multirow{2}{*}{\textbf{Model}} & \multirow{2}{*}{\textbf{Method}} & \multicolumn{2}{c|}{\textbf{General}} & \multicolumn{2}{c|}{\textbf{Math (acc)}} & \multicolumn{2}{c|}{\textbf{Code (pass@1)}} & \multirow{2}{*}{\textbf{Avg.}} \\ \cline{3-8}
 &  & \textbf{AlpacaEval} & \textbf{MMLU} & \textbf{GSM8K} & \textbf{MATH} & \textbf{HumanEval} & \textbf{MBPP} &  \\ \hline
LM & - & 82.72 & 53.34 & 45.79 & 0.14 & 30.48 & 31.40 & 40.65 \\
Math & - & - & - & 63.08 & 11.60 & - & - & - \\
Code & - & - & - & - & - & 23.78 & 27.20 & - \\ \hline
\multirow{6}{*}{\begin{tabular}[c]{@{}l@{}}LM \\ + Math \\     + Code\end{tabular}} 
 & TA & 78.93 & 51.04 & 58.45 & 9.88 & 18.29 & 29.80 & 41.07 \\
 & TIES & 80.53 & 54.30 & 62.55 & 9.54 & 21.95 & 30.40 & 43.21 \\
 & DARE & 75.00 & 54.12 & 58.00 & 9.20 & \textbf{29.27} & \textbf{31.40} & 42.83 \\
 & DELLA & \textbf{83.16} & 53.52 & 61.80 & 7.88 & 19.50 & \textbf{31.40} & 42.87 \\
 & TALL-Mask & 80.31 & 54.25 & 62.70 & 10.62 & 20.73 & 30.80 & 43.23 \\
 & PCB & 81.98 & 53.37 & 63.83 & 8.24 & 26.22 & 26.60 & 43.37 \\
 & \textbf{OBIM} & 81.23 & \textbf{54.39} & \textbf{68.23} & \textbf{12.50} & 25.61 & 29.40 & \textbf{45.23} \\ \hline
\end{tabular}%
}
\caption{Performance comparison of SFT model merging. The results for each individual model are presented at the top of the table, the lower section displays the results of merging the three models using different methods.}
\label{tab:main_result}
\end{table*}

We conduct experiments on both SFT models and post-pretrained checkpoints to demonstrate the effectiveness of our method. To validate its robustness, we evaluate OBIM using two popular backbone models, LLaMA2 \cite{llama2} and Qwen2 \cite{qwen2}, in separate experiments. We also perform an ablation study to analyze the contributions of specific components within OBIM. Furthermore, we investigate key factors in our method to assess their influence on the final performance.

\subsection{Experimental Setup}

\paragraph{Experiment Settings for SFT Models.} 
Following previous works \cite{dare, della}, we use \textit{Llama-2-13b} as the pre-trained backbone and incorporate three fine-tuned models for cross-task merging experiments: \textit{WizardLM-13B-V1.2} \cite{wizardlm} for instruction following, \textit{WizardMath-13B-V1.0} \cite{wizardmath} for mathematical reasoning, and \textit{llama-2-13b-code-alpaca} \cite{llamacode} for code generation. To evaluate the capabilities of the merged models, we use AlpacaEval \cite{alpacaeval} and MMLU \cite{mmlu} for general understanding; GSM8K \cite{gsm8k} and MATH \cite{math} for mathematical reasoning; and HumanEval \cite{humaneval} and MBPP \cite{mbpp} for code generation. Performance is measured using the win rate for AlpacaEval\footnote{We calculated the win rate by comparing target model to \textit{text-davinci-003}, using \textit{GPT-4o} as the evaluator.}, zero-shot accuracy for MMLU, GSM8K, and MATH, and pass@1 for HumanEval and MBPP.


\paragraph{Experiment Settings for Post-pretrained Models.}
We perform post-pretraining on \textit{Qwen2-7B} using a multilingual dataset to enhance its Japanese proficiency. The dataset consists of over 200 billion tokens and includes publicly available pretraining corpora in English, Chinese, and Japanese. Details of the dataset are provided in Appendix~\ref{appendix:pt_data}. The three best-performing checkpoints on Japanese evaluation are selected and merged with the backbone model. To assess performance across different languages, we employ three benchmarks: C-Eval \cite{ceval} for Chinese, MMLU for English, and the Japanese Language Model Evaluation Harness (JP-LMEH) \footnote{https://github.com/Stability-AI/lm-evaluation-harness/tree/jp-stable} for Japanese. Five-shot accuracy is used for evaluating C-Eval and MMLU. JP-LMEH encompasses nine distinct NLP tasks\footnote{JSQuAD, JCommonsenseQA, JNLI, MARC-ja, XLSum-ja, JCoLA, MGSM-ja, XWinograd-ja, and JAQKET}, with the average score serving as an indicator of overall Japanese language proficiency.

\paragraph{Baselines.}
We compare our method with the following state-of-the-art model merging approaches applicable to LLMs:  
\textbf{TA} \cite{ta}: Task Arithmetic, a simple delta weight merging method that does not explicitly address interference.  
\textbf{TIES} \cite{ties}: Eliminates redundant parameters based on magnitude and resolves sign conflicts.  
\textbf{DARE} \cite{dare}: Drop And REscale, Randomly drops a proportion of parameters and rescales the remaining ones to reduce redundancy.  
\textbf{DELLA} \cite{della}: Assigns dropout probabilities to parameters based on their magnitudes for pruning.  
\textbf{TALL-Mask} \cite{della}: Uses a masking mechanism to filter out parameters relevant to only a few tasks.  
\textbf{PCB} \cite{pcb}: Leverages parameter competition to optimize the merging process.  

\paragraph{Validation Set.}
Since OBIM requires a small sample set for parameter saliency computation, we hold out a validation set comprising portions of the training and development sets from each benchmark. Specifically, we compute saliency using data related to the model's training task: AlpacaEval and MMLU for general models, GSM8K and MATH for math models, and MBPP for code models. For post-pretrained models, saliency is computed using a multilingual dataset consisting of C-Eval, MMLU, and JP-LMEH. Details are provided in Appendix~\ref{appendix:valid_set}.

\subsection{Main Results}
\paragraph{Results on SFT Models.}
The results are summarized in Table \ref{tab:main_result}. We first present the performance of each individual model, followed by the results of merging the three task-specific experts using different methods. As shown in Table \ref{tab:main_result}, OBIM achieves significant improvements in mathematical reasoning, with a $5.15\%$ gain on GSM8K and a $0.9\%$ gain on MATH compared to the original math model. In contrast, many other methods fail to surpass the source math model. For other benchmarks, OBIM ranks first on MMLU and remains competitive across other tasks. However, its performance on code generation is relatively lower. We suspect this is because the general model, rather than the code model, performs best on code generation, yet its saliency is computed using general data, leading to suboptimal preservation of coding ability. Overall, OBIM achieves the highest average performance, outperforming the second-best method by $1.86\%$, demonstrating its effectiveness in merging models for task fusion.

\begin{table}[htbp]
\renewcommand{\arraystretch}{1.05}
\centering
\resizebox{1.02\columnwidth}{!}{%
\begin{tabular}{l|c|ccc|c}
\hline
\textbf{Model} & \textbf{Method} & \textbf{C-Eval} & \textbf{MMLU} & \textbf{JP-LMEH} & \textbf{Avg.} \\ \hline
Qwen2-7B & - & 83.51 & 69.22 & 69.19 & 73.97 \\
ckpt-1 & - & 77.71 & 67.01 & 72.13 & 72.28 \\
ckpt-2 & - & 75.48 & 67.05 & 71.69 & 71.41 \\ 
ckpt-3 & - & 74.37 & 66.86 & 71.61 & 70.95 \\ \hline
\multirow{6}{*}{\begin{tabular}[c]{@{}l@{}} Qwen2-7B \\ + ckpt-1 \\ + ckpt-2 \\ + ckpt-3 \end{tabular}} & TA & 76.15 & 68.47 & 71.90 & 72.17 \\
 & TIES & 76.52 & 67.92 & 71.93 & 72.12 \\
 & DARE & 78.97 & 68.85 & 71.21 & 73.01 \\
 & DELLA & 77.26 & 68.51 & 72.05 & 72.61 \\
 & TALL-Mask & 77.72 & 68.48 & 71.56 & 72.59 \\
 & PCB & 77.71 & 68.31 & 72.08 & 72.70 \\
 & \textbf{OBIM} & \textbf{80.46} & \textbf{69.89} & \textbf{72.14} & \textbf{74.16} \\ \hline
\end{tabular}%
}
\caption{Performance comparison of post-pretrained model merging. \textit{ckpt-1}, \textit{ckpt-2} and \textit{ckpt-3} are the checkpoints that achieve the best JP-LMEH results during post-pretraining but exhibit performance degradation in Chinese and English.}
\label{tab:multi-ckpt}
\end{table}

\begin{figure}[ht]
\centering 
\includegraphics[width=1\columnwidth]{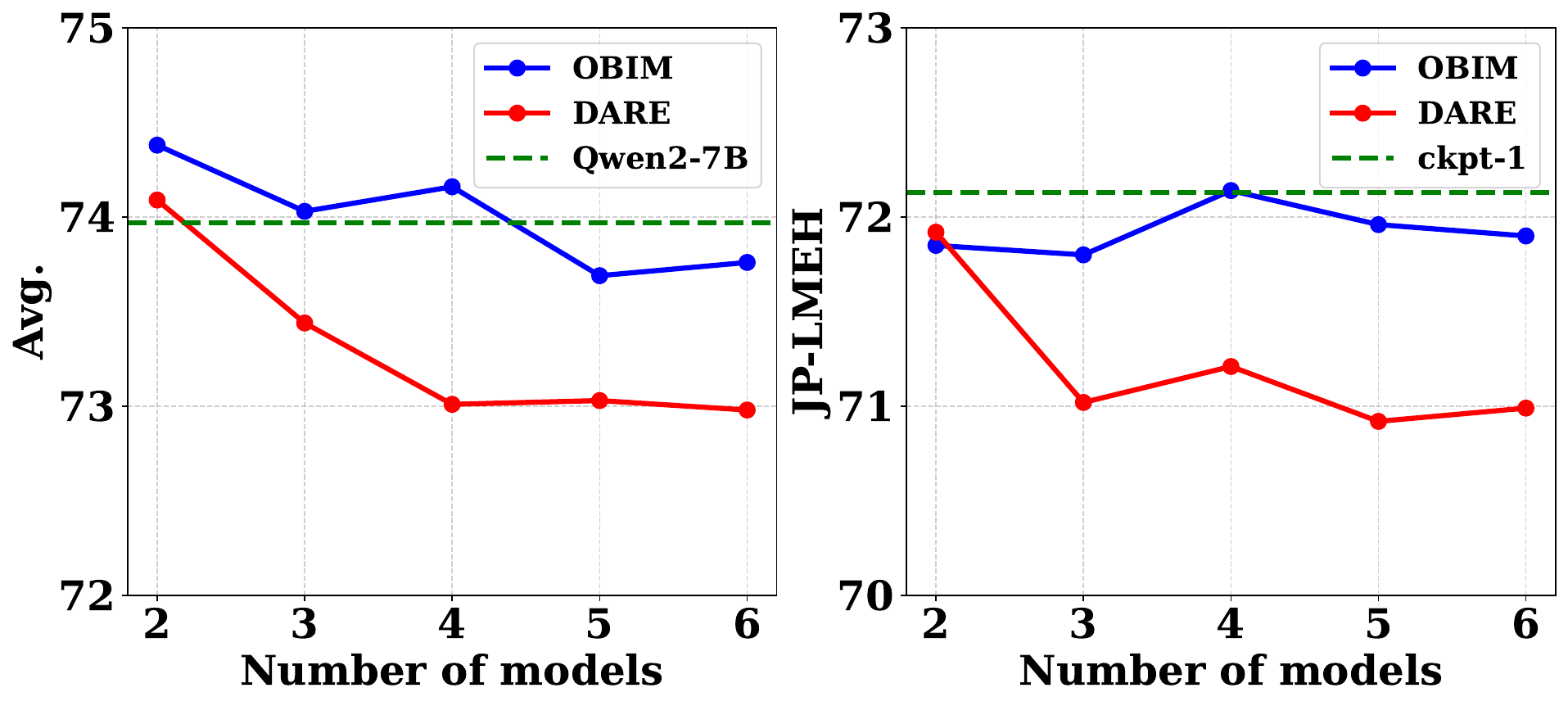}
\caption{Performance comparison of merging different numbers of models between OBIM and DARE. The left part presents the average performance across three languages, while the right part shows the results for Japanese capability. The green dotted line represents the best performance of models before merging.}
\label{fig:exp} 
\end{figure}

\begin{table*}[htbp]
\renewcommand{\arraystretch}{1.1}
\centering
\resizebox{\textwidth}{!}{%
\begin{tabular}{c|c|c|cc|cc|cc|c}
\hline
\multirow{2}{*}{\textbf{Method}} & \multirow{2}{*}{\textbf{Intra-model}} & \multirow{2}{*}{\textbf{Inter-model}} & \multicolumn{2}{c|}{\textbf{General}} & \multicolumn{2}{c|}{\textbf{Math (acc)}} & \multicolumn{2}{c|}{\textbf{Code (pass@1)}} & \multirow{2}{*}{\textbf{Avg.}} \\ \cline{4-9}
 &  &  & \textbf{AlpacaEval} & \textbf{MMLU} & \textbf{GSM8K} & \textbf{MATH} & \textbf{HumanEval} & \textbf{MBPP} &  \\ \hline
TIES & Magnitude & \multirow{2}{*}{Disjoint Mean} & 80.53 & 54.30 & 62.55 & 9.54 & 21.95 & 30.40 & 43.21 \\
TIES+OBM & Saliency &  & 79.03 & \textbf{54.44} & 64.29 & 10.52 & \textbf{26.83} & \textbf{30.80} & 44.32 \\ \hline
TIES+IM & Magnitude & \multirow{2}{*}{Iterative merging} & 80.81 & 54.31 & 68.08 & \textbf{13.08} & 21.95 & 30.40 & 44.77 \\
\textbf{OBIM} & Saliency &  & \textbf{81.23} & 54.39 & \textbf{68.23} & 12.50 & 25.61 & 29.40 & \textbf{45.23} \\ \hline
\end{tabular}%
}
\caption{Ablation study on components in OBIM. Each method's strategy for mitigating intra-model and inter-model interference is listed in the columns \textit{Intra-model} and \textit{Inter-model}, respectively.}
\label{tab:ablation_study}
\end{table*}

\paragraph{Results on Post-pretrained Models.}
As shown in Table \ref{tab:multi-ckpt}, checkpoints trained from \textit{Qwen2-7B} exhibit enhanced Japanese proficiency but experience a significant decline in the Chinese and English capabilities of the base model. The goal of model merging is to restore Chinese and English performance to the base model level while preserving the improved Japanese ability of the continually trained model.
Our results show that while no merging method fully restores C-Eval performance to the base model level, OBIM achieves the highest recovery rate, surpassing other methods by $1.49\%$. For English and Japanese, OBIM is the only method that surpasses the base model on MMLU and JP-LMEH, achieving improvements of $1.04\%$ and $0.04\%$ over other approaches. In terms of overall multilingual performance, OBIM ranks first across all benchmarks, producing a model with the strongest overall capabilities.

To further validate the robustness of our method, we evaluate the performance of merging different numbers of models. Specifically, we select the five best checkpoints during post-pretraining and merge them with the base model. We compare our method with DARE, as well as the best-performing individual models before merging, and present the results in Figure~\ref{fig:exp}\footnote{Detailed results are provided in Appendix~\ref{appendix:multi}}.  
The results indicate that performance declines as the number of merged models increases, likely due to increased interference among models. However, our method remains stable and achieves performance competitive with the best individual model in both the average and Japanese capability evaluations. In contrast, DARE underperforms by approximately $1\%$, demonstrating that our method more effectively mitigates interference when merging multiple models.

\subsection{Ablation Study}
\label{sec:ablationstudy}
To assess the contributions of the two key components, OBM and IM, we conduct experiments using SFT model merging settings. Since OBM and IM cannot perform merging independently, we use TIES as the baseline method and integrate OBM and IM with the weight consensus approach for resolving sign conflict, termed "Disjoint Mean," as well as the magnitude-based parameter trimming method in TIES, respectively. Details are provided in Appendix~\ref{appendix:ablation}.

The results are presented in Table~\ref{tab:ablation_study}, where we also outline each method’s strategy for mitigating intra-model and inter-model interference. By comparing TIES with TIES+OBM and TIES+IM with OBIM, where each pair employs the same method for reducing inter-model interference, we observe that methods utilizing saliency-based parameter selection outperform those relying on magnitude-based selection. This finding confirms the superiority of OBM. Furthermore, methods incorporating IM consistently outperform their counterparts using the same intra-model approach. Specifically, TIES+IM surpasses TIES, and OBIM outperforms TIES+OBM, demonstrating the effectiveness of IM. In general, OBIM achieves the highest performance among all methods, proving that the combination of OBM and IM can further enhance the performance.

\subsection{Influence of the Validation Set}  
\label{sec:exp_val}
Samples in the validation set influence saliency scores. To assess this impact, we replace the samples used for computing the saliency of the math model under the experimental settings of SFT model merging and evaluate the merging performance on two mathematical tasks. 

As shown in Table~\ref{tab:data_selection}, we compare the results using task-related data from GSM8K and MATH (\textit{Math}), irrelevant data from MBPP (\textit{Code}), and a mixed dataset (\textit{Math+Code}). All datasets contain the same number of samples. \textit{Math} achieves the best performance, followed by \textit{Math+Code}, while \textit{Code} performs the worst. This suggests that using task-specific data for saliency computation enhances task knowledge retention. We attribute this to our assumption for saliency approximation, which requires the first derivative to be approximately zero. This condition implies that data well learned by the model enables more accurate saliency estimation.

\begin{table}[htbp]
\centering
\resizebox{0.95\columnwidth}{!}{%
\begin{tabular}{c|c|cc|c}
\hline
\multirow{2}{*}{\textbf{Source}} & \multirow{2}{*}{\textbf{Size}} & \multicolumn{2}{c|}{\textbf{Math (acc)}} & \multirow{2}{*}{\textbf{Avg.}} \\ \cline{3-4}
 &  & \textbf{GSM8K} & \textbf{MATH} &  \\ \hline
Math & \multirow{3}{*}{100} & \textbf{68.23} & \textbf{12.50} & \textbf{40.37} \\
Code &  & 67.48 & 11.84 & 39.66 \\
Math + Code &  & {\ul 68.01} & {\ul 12.42} & {\ul 40.22} \\ \hline
\end{tabular}%
}
\caption{Comparison of different validation sets used for saliency computation. The winners and runners-up are marked in bold font and underlined, respectively.}
\label{tab:data_selection}
\end{table}

We also investigate the impact of sample size by using different numbers of samples from the same source. The results in Table~\ref{tab:data_amount} indicate that 100 samples yield the best performance, while using either more or fewer samples leads to a decline. However, the results with only 10 samples remain competitive, suggesting that the method is effective even with a limited number of samples.

\begin{table}[htbp]
\centering
\resizebox{0.85\columnwidth}{!}{%
\begin{tabular}{c|c|cc|c}
\hline
\multirow{2}{*}{\textbf{Source}} & \multirow{2}{*}{\textbf{Size}} & \multicolumn{2}{c|}{\textbf{Math (acc)}} & \multirow{2}{*}{\textbf{Avg.}} \\ \cline{3-4}
 &  & \textbf{GSM8K} & \textbf{MATH} &  \\ \hline
\multirow{4}{*}{Math} & 10 & {\ul 68.16} & 12.10 & {\ul 40.13} \\
 & 50 & 67.40 & 11.88 & 39.64 \\
 & 100 & \textbf{68.23} & \textbf{12.50} & \textbf{40.37} \\
 & 200 & 67.70 & {\ul 12.28} & 39.99 \\ \hline
\end{tabular}%
}
\caption{Performance across different sample sizes. 100 samples achieve the highest performance, followed by 10 samples.}
\label{tab:data_amount}
\end{table}

\subsection{Influence of Iterative Merging Order}  
\label{sec:exp_order}
We analyze how the order of iterative merging influences performance. We conduct experiments based on the SFT model merging settings with four different merging orders: shifting the order using a rotation operation across model layers (\textit{Rotation}), prioritizing the math model (\textit{Math First}), placing the math model last (\textit{Math Last}), and prioritizing the general language model (\textit{LM First}). We evaluate the results on mathematical tasks.

As shown in Table~\ref{tab:order}, \textit{Math First} achieves the best performance, whereas \textit{Math Last} performs the worst, and \textit{LM First} also yields poor results. These findings suggest that the earlier a model is merged, the better its knowledge is preserved. However, while \textit{Rotation} performs slightly worse than \textit{Math First}, it still surpasses the original math model, demonstrating its robustness. We attribute this to the redundancy of parameters in LLMs, where core capabilities can be largely retained even when only a subset of layers is preserved.

\begin{table}[htbp]
\centering
\resizebox{0.85\columnwidth}{!}{%
\begin{tabular}{c|cc|c}
\hline
\multirow{2}{*}{\textbf{Iteration Order}} & \multicolumn{2}{c|}{\textbf{Math (acc)}} & \multirow{2}{*}{\textbf{Avg.}} \\ \cline{2-3}
 & \textbf{GSM8K} & \textbf{MATH} &  \\ \hline
Rotation & {\ul 67.10} & {\ul 11.96} & {\ul 39.53} \\
Math First & \textbf{67.48} & \textbf{14.38} & \textbf{40.93} \\
Math Last & 61.71 & 2.06 & 31.89 \\
LM First & 63.15 & 3.14 & 33.15 \\ \hline
\end{tabular}%
}
\caption{Performance comparison of different merging orders in iterative merging.}
\label{tab:order}
\end{table}

\section{Related Work}

Model merging has gained popularity in LLM research \cite{metagpt, yang2024model}. By amalgamating multiple homologous LLMs into a single model, this technique has been applied to address several challenges, such as building multi-task experts \cite{cai2023robust, fusechat}, detoxification \cite{hu2024separate, zhang2023composing}, and preference alignment \cite{dogerm, rame2024rewarded}. Model merging methods are primarily based on two fundamental approaches: weight averaging \cite{modelsoups} and task arithmetic \cite{ta}. 

Weight-based model merging methods design rules or matrices to determine merging coefficients. For example, RegMean \cite{jin2022dataless} optimizes a linear regression problem for linear weights, Fisher-Merging \cite{matena2022merging} uses the Fisher information matrix to assess parameter importance. Some works explore the space of these coefficients using parameter searching algorithms, such as evolutionary algorithms \cite{evolutionary} or Bayesian optimization \cite{liu2024checkpoint}. Although these methods demonstrate effectiveness, they suffer from inefficiency: parameter search is time-consuming, and solving the objectives requires substantial computation resources. 

Subspace-based model merging methods focus on eliminating insignificant parameters and merging sparse models within the parameter subspace to reduce interference. TIES \cite{ties} trims individual models based on parameter magnitudes, while Model Breadcrumbs \cite{model-breadcrumbs} refines this by removing both low-magnitude and high-magnitude outliers. DARE \cite{dare} emphasizes the importance of rescaling after sparsification, and TALL-Mask \cite{tallmask} creates task-specific mask matrices based on predefined thresholds to filter out irrelevant parameters. However, these methods are limited to specific patterns, such as sign conflicts or threshold-based filtering, and magnitude-based sparsification remains suboptimal. To better address the interference problem, we propose a solution based on parameter saliency sparsification and a mutually exclusive iterative merging framework.

\section{Conclusion}


In this work, we propose OBIM, a novel merging method for LLMs that selectively retains representative delta weights based on saliency and iteratively integrates task vectors to reduce both intra-model and inter-model interference. OBIM achieves state-of-the-art performance in merging SFT models and post-pretraining checkpoints, demonstrating its effectiveness and versatility. Extensive ablation studies further validate its key components. Additionally, OBIM is computationally efficient and memory-light, making it well-suited for real-world applications.

\section{Limitations}

While our work provides valuable insights into LLM merging, several limitations should be noted:  
(1) The application of OBIM relies on models with identical architectures and shared initializations, limiting its applicability to diverse model types.  
(2) Although efficient, OBIM requires an additional validation set and incurs extra computational costs for saliency computation compared to magnitude-based methods. 
(3) Our analysis primarily focuses on interference from the perspective of parameter aggregation, with limited theoretical exploration, highlighting the need for further research in future work.

\bibliography{acl_latex}

\begin{thebibliography}{47}
\providecommand{\natexlab}[1]{#1}

\bibitem[{Akiba et~al.(2025)Akiba, Shing, Tang, Sun, and Ha}]{evolutionary}
Takuya Akiba, Makoto Shing, Yujin Tang, Qi~Sun, and David Ha. 2025.
\newblock Evolutionary optimization of model merging recipes.
\newblock \emph{Nature Machine Intelligence}, pages 1--10.

\bibitem[{Austin et~al.(2021)Austin, Odena, Nye, Bosma, Michalewski, Dohan, Jiang, Cai, Terry, Le et~al.}]{mbpp}
Jacob Austin, Augustus Odena, Maxwell Nye, Maarten Bosma, Henryk Michalewski, David Dohan, Ellen Jiang, Carrie Cai, Michael Terry, Quoc Le, et~al. 2021.
\newblock Program synthesis with large language models.
\newblock \emph{arXiv preprint arXiv:2108.07732}.

\bibitem[{Bowen et~al.(2024)Bowen, Songning, Jiemin, Zhihao, Shiming, and Yutao}]{bowen2024beyond}
Tian Bowen, Lai Songning, Wu~Jiemin, Shuai Zhihao, Ge~Shiming, and Yue Yutao. 2024.
\newblock Beyond task vectors: Selective task arithmetic based on importance metrics.
\newblock \emph{arXiv preprint arXiv:2411.16139}.

\bibitem[{Cai et~al.(2023)Cai, Zhang, and Wang}]{cai2023robust}
Ruisi Cai, Zhenyu Zhang, and Zhangyang Wang. 2023.
\newblock Robust weight signatures: gaining robustness as easy as patching weights?
\newblock In \emph{International Conference on Machine Learning}, pages 3495--3506. PMLR.

\bibitem[{Chaudhary(2023)}]{llamacode}
Sahil Chaudhary. 2023.
\newblock Code alpaca: An instruction-following llama model for code generation.
\newblock \url{https://github.com/sahil280114/codealpaca}.

\bibitem[{Chen et~al.(2021)Chen, Tworek, Jun, Yuan, Pinto, Kaplan, Edwards, Burda, Joseph, Brockman et~al.}]{humaneval}
Mark Chen, Jerry Tworek, Heewoo Jun, Qiming Yuan, Henrique Ponde De~Oliveira Pinto, Jared Kaplan, Harri Edwards, Yuri Burda, Nicholas Joseph, Greg Brockman, et~al. 2021.
\newblock Evaluating large language models trained on code.
\newblock \emph{arXiv preprint arXiv:2107.03374}.

\bibitem[{Choudhary et~al.(2020)Choudhary, Mishra, Goswami, and Sarangapani}]{choudhary2020comprehensive}
Tejalal Choudhary, Vipul Mishra, Anurag Goswami, and Jagannathan Sarangapani. 2020.
\newblock A comprehensive survey on model compression and acceleration.
\newblock \emph{Artificial Intelligence Review}, 53:5113--5155.

\bibitem[{Cobbe et~al.(2021)Cobbe, Kosaraju, Bavarian, Chen, Jun, Kaiser, Plappert, Tworek, Hilton, Nakano et~al.}]{gsm8k}
Karl Cobbe, Vineet Kosaraju, Mohammad Bavarian, Mark Chen, Heewoo Jun, Lukasz Kaiser, Matthias Plappert, Jerry Tworek, Jacob Hilton, Reiichiro Nakano, et~al. 2021.
\newblock Training verifiers to solve math word problems.
\newblock \emph{arXiv preprint arXiv:2110.14168}.

\bibitem[{Davari and Belilovsky(2025)}]{model-breadcrumbs}
MohammadReza Davari and Eugene Belilovsky. 2025.
\newblock Model breadcrumbs: Scaling multi-task model merging with sparse masks.
\newblock In \emph{European Conference on Computer Vision}, pages 270--287. Springer.

\bibitem[{Deep et~al.(2024)Deep, Bhardwaj, and Poria}]{della}
Pala~Tej Deep, Rishabh Bhardwaj, and Soujanya Poria. 2024.
\newblock Della-merging: Reducing interference in model merging through magnitude-based sampling.
\newblock \emph{arXiv preprint arXiv:2406.11617}.

\bibitem[{Dekoninck et~al.(2023)Dekoninck, Fischer, Beurer-Kellner, and Vechev}]{controlled}
Jasper Dekoninck, Marc Fischer, Luca Beurer-Kellner, and Martin Vechev. 2023.
\newblock Controlled text generation via language model arithmetic.
\newblock \emph{arXiv preprint arXiv:2311.14479}.

\bibitem[{DU et~al.(2024)DU, Lee, Li, Jiang, Guo, Yu, Liu, Goh, Tang, He, and Zhang}]{pcb}
Guodong DU, Junlin Lee, Jing Li, Runhua Jiang, Yifei Guo, Shuyang Yu, Hanting Liu, Sim~Kuan Goh, Ho-Kin Tang, Daojing He, and Min Zhang. 2024.
\newblock \href {https://openreview.net/forum?id=l5SbrtvSRS} {Parameter competition balancing for model merging}.
\newblock In \emph{The Thirty-eighth Annual Conference on Neural Information Processing Systems}.

\bibitem[{Frantar and Alistarh(2022)}]{obc}
Elias Frantar and Dan Alistarh. 2022.
\newblock Optimal brain compression: A framework for accurate post-training quantization and pruning.
\newblock \emph{Advances in Neural Information Processing Systems}, 35:4475--4488.

\bibitem[{Frantar et~al.(2022)Frantar, Ashkboos, Hoefler, and Alistarh}]{gptq}
Elias Frantar, Saleh Ashkboos, Torsten Hoefler, and Dan Alistarh. 2022.
\newblock Gptq: Accurate post-training quantization for generative pre-trained transformers.
\newblock \emph{arXiv preprint arXiv:2210.17323}.

\bibitem[{Hassibi et~al.(1993)Hassibi, Stork, and Wolff}]{obs}
Babak Hassibi, David~G Stork, and Gregory~J Wolff. 1993.
\newblock Optimal brain surgeon and general network pruning.
\newblock In \emph{IEEE international conference on neural networks}, pages 293--299. IEEE.

\bibitem[{He and Xiao(2023)}]{he2023structured}
Yang He and Lingao Xiao. 2023.
\newblock Structured pruning for deep convolutional neural networks: A survey.
\newblock \emph{IEEE transactions on pattern analysis and machine intelligence}.

\bibitem[{Hendrycks et~al.(2021{\natexlab{a}})Hendrycks, Burns, Basart, Zou, Mazeika, Song, and Steinhardt}]{mmlu}
Dan Hendrycks, Collin Burns, Steven Basart, Andy Zou, Mantas Mazeika, Dawn Song, and Jacob Steinhardt. 2021{\natexlab{a}}.
\newblock \href {https://openreview.net/forum?id=d7KBjmI3GmQ} {Measuring massive multitask language understanding}.
\newblock In \emph{International Conference on Learning Representations}.

\bibitem[{Hendrycks et~al.(2021{\natexlab{b}})Hendrycks, Burns, Kadavath, Arora, Basart, Tang, Song, and Steinhardt}]{math}
Dan Hendrycks, Collin Burns, Saurav Kadavath, Akul Arora, Steven Basart, Eric Tang, Dawn Song, and Jacob Steinhardt. 2021{\natexlab{b}}.
\newblock Measuring mathematical problem solving with the math dataset.
\newblock \emph{NeurIPS}.

\bibitem[{Hu et~al.(2024)Hu, Li, Hu, Zheng, Liu, and Zhang}]{hu2024separate}
Xinshuo Hu, Dongfang Li, Baotian Hu, Zihao Zheng, Zhenyu Liu, and Min Zhang. 2024.
\newblock Separate the wheat from the chaff: Model deficiency unlearning via parameter-efficient module operation.
\newblock In \emph{Proceedings of the AAAI Conference on Artificial Intelligence}, volume~38, pages 18252--18260.

\bibitem[{Huang et~al.(2023)Huang, Bai, Zhu, Zhang, Zhang, Su, Liu, Lv, Zhang, Lei, Fu, Sun, and He}]{ceval}
Yuzhen Huang, Yuzhuo Bai, Zhihao Zhu, Junlei Zhang, Jinghan Zhang, Tangjun Su, Junteng Liu, Chuancheng Lv, Yikai Zhang, Jiayi Lei, Yao Fu, Maosong Sun, and Junxian He. 2023.
\newblock C-eval: A multi-level multi-discipline chinese evaluation suite for foundation models.
\newblock In \emph{Advances in Neural Information Processing Systems}.

\bibitem[{Ilharco et~al.(2023)Ilharco, Ribeiro, Wortsman, Schmidt, Hajishirzi, and Farhadi}]{ta}
Gabriel Ilharco, Marco~Tulio Ribeiro, Mitchell Wortsman, Ludwig Schmidt, Hannaneh Hajishirzi, and Ali Farhadi. 2023.
\newblock \href {https://openreview.net/forum?id=6t0Kwf8-jrj} {Editing models with task arithmetic}.
\newblock In \emph{The Eleventh International Conference on Learning Representations}.

\bibitem[{Jang et~al.(2024)Jang, Yun, and Han}]{jang2024model}
Dong-Hwan Jang, Sangdoo Yun, and Dongyoon Han. 2024.
\newblock Model stock: All we need is just a few fine-tuned models.
\newblock In \emph{European Conference on Computer Vision}, pages 207--223. Springer.

\bibitem[{Jin et~al.(2022)Jin, Ren, Preotiuc-Pietro, and Cheng}]{jin2022dataless}
Xisen Jin, Xiang Ren, Daniel Preotiuc-Pietro, and Pengxiang Cheng. 2022.
\newblock Dataless knowledge fusion by merging weights of language models.
\newblock \emph{arXiv preprint arXiv:2212.09849}.

\bibitem[{Kim et~al.(2024)Kim, Kim, Kim, Castells, Choi, Shin, and Song}]{kim2024shortened}
Bo-Kyeong Kim, Geonmin Kim, Tae-Ho Kim, Thibault Castells, Shinkook Choi, Junho Shin, and Hyoung-Kyu Song. 2024.
\newblock Shortened llama: A simple depth pruning for large language models.
\newblock \emph{arXiv preprint arXiv:2402.02834}, 11.

\bibitem[{LeCun et~al.(1989)LeCun, Denker, and Solla}]{obd}
Yann LeCun, John Denker, and Sara Solla. 1989.
\newblock Optimal brain damage.
\newblock \emph{Advances in neural information processing systems}, 2.

\bibitem[{Li et~al.(2024)Li, Fang, Smyrnis, Ivgi, Jordan, Gadre, Bansal, Guha, Keh, Arora et~al.}]{li2024datacomp}
Jeffrey Li, Alex Fang, Georgios Smyrnis, Maor Ivgi, Matt Jordan, Samir Gadre, Hritik Bansal, Etash Guha, Sedrick Keh, Kushal Arora, et~al. 2024.
\newblock Datacomp-lm: In search of the next generation of training sets for language models.
\newblock \emph{arXiv preprint arXiv:2406.11794}.

\bibitem[{Li et~al.(2023)Li, Zhang, Dubois, Taori, Gulrajani, Guestrin, Liang, and Hashimoto}]{alpacaeval}
Xuechen Li, Tianyi Zhang, Yann Dubois, Rohan Taori, Ishaan Gulrajani, Carlos Guestrin, Percy Liang, and Tatsunori~B. Hashimoto. 2023.
\newblock Alpacaeval: An automatic evaluator of instruction-following models.
\newblock \url{https://github.com/tatsu-lab/alpaca_eval}.

\bibitem[{Lin et~al.(2024)Lin, Li, Lee, and Chen}]{dogerm}
Tzu-Han Lin, Chen-An Li, Hung-yi Lee, and Yun-Nung Chen. 2024.
\newblock Dogerm: Equipping reward models with domain knowledge through model merging.
\newblock \emph{arXiv preprint arXiv:2407.01470}.

\bibitem[{Liu et~al.(2024)Liu, Wang, Wang, Chen, Li, Tu, Chu, Li, and Sui}]{liu2024checkpoint}
Deyuan Liu, Zecheng Wang, Bingning Wang, Weipeng Chen, Chunshan Li, Zhiying Tu, Dianhui Chu, Bo~Li, and Dianbo Sui. 2024.
\newblock Checkpoint merging via bayesian optimization in llm pretraining.
\newblock \emph{arXiv preprint arXiv:2403.19390}.

\bibitem[{Luo et~al.(2023)Luo, Sun, Xu, Zhao, Lou, Tao, Geng, Lin, Chen, and Zhang}]{wizardmath}
Haipeng Luo, Qingfeng Sun, Can Xu, Pu~Zhao, Jianguang Lou, Chongyang Tao, Xiubo Geng, Qingwei Lin, Shifeng Chen, and Dongmei Zhang. 2023.
\newblock Wizardmath: Empowering mathematical reasoning for large language models via reinforced evol-instruct.
\newblock \emph{arXiv preprint arXiv:2308.09583}.

\bibitem[{Matena and Raffel(2022)}]{matena2022merging}
Michael~S Matena and Colin~A Raffel. 2022.
\newblock Merging models with fisher-weighted averaging.
\newblock \emph{Advances in Neural Information Processing Systems}, 35:17703--17716.

\bibitem[{Rame et~al.(2024)Rame, Couairon, Dancette, Gaya, Shukor, Soulier, and Cord}]{rame2024rewarded}
Alexandre Rame, Guillaume Couairon, Corentin Dancette, Jean-Baptiste Gaya, Mustafa Shukor, Laure Soulier, and Matthieu Cord. 2024.
\newblock Rewarded soups: towards pareto-optimal alignment by interpolating weights fine-tuned on diverse rewards.
\newblock \emph{Advances in Neural Information Processing Systems}, 36.

\bibitem[{Shoemake(1985)}]{slerp}
Ken Shoemake. 1985.
\newblock Animating rotation with quaternion curves.
\newblock In \emph{Proceedings of the 12th annual conference on Computer graphics and interactive techniques}, pages 245--254.

\bibitem[{Sun et~al.(2023)Sun, Liu, Bair, and Kolter}]{sun2023simple}
Mingjie Sun, Zhuang Liu, Anna Bair, and J~Zico Kolter. 2023.
\newblock A simple and effective pruning approach for large language models.
\newblock \emph{arXiv preprint arXiv:2306.11695}.

\bibitem[{Touvron et~al.(2023)Touvron, Martin, Stone, Albert, Almahairi, Babaei, Bashlykov, Batra, Bhargava, Bhosale et~al.}]{llama2}
Hugo Touvron, Louis Martin, Kevin Stone, Peter Albert, Amjad Almahairi, Yasmine Babaei, Nikolay Bashlykov, Soumya Batra, Prajjwal Bhargava, Shruti Bhosale, et~al. 2023.
\newblock Llama 2: Open foundation and fine-tuned chat models.
\newblock \emph{arXiv preprint arXiv:2307.09288}.

\bibitem[{Wan et~al.(2024{\natexlab{a}})Wan, Huang, Cai, Quan, Bi, and Shi}]{wan2024knowledge}
Fanqi Wan, Xinting Huang, Deng Cai, Xiaojun Quan, Wei Bi, and Shuming Shi. 2024{\natexlab{a}}.
\newblock Knowledge fusion of large language models.
\newblock \emph{arXiv preprint arXiv:2401.10491}.

\bibitem[{Wan et~al.(2024{\natexlab{b}})Wan, Zhong, Yang, Chen, and Quan}]{fusechat}
Fanqi Wan, Longguang Zhong, Ziyi Yang, Ruijun Chen, and Xiaojun Quan. 2024{\natexlab{b}}.
\newblock Fusechat: Knowledge fusion of chat models.
\newblock \emph{arXiv preprint arXiv:2408.07990}.

\bibitem[{Wang et~al.(2024)Wang, Dimitriadis, Ortiz{-}Jim{\'{e}}nez, Fleuret, and Frossard}]{tallmask}
Ke~Wang, Nikolaos Dimitriadis, Guillermo Ortiz{-}Jim{\'{e}}nez, Fran\c{c}ois Fleuret, and Pascal Frossard. 2024.
\newblock Localizing task information for improved model merging and compression.
\newblock In \emph{International Conference on Machine Learning}.

\bibitem[{Wortsman et~al.(2022)Wortsman, Ilharco, Gadre, Roelofs, Gontijo-Lopes, Morcos, Namkoong, Farhadi, Carmon, Kornblith et~al.}]{modelsoups}
Mitchell Wortsman, Gabriel Ilharco, Samir~Ya Gadre, Rebecca Roelofs, Raphael Gontijo-Lopes, Ari~S Morcos, Hongseok Namkoong, Ali Farhadi, Yair Carmon, Simon Kornblith, et~al. 2022.
\newblock Model soups: averaging weights of multiple fine-tuned models improves accuracy without increasing inference time.
\newblock In \emph{International conference on machine learning}, pages 23965--23998. PMLR.

\bibitem[{Xu et~al.(2023)Xu, Sun, Zheng, Geng, Zhao, Feng, Tao, and Jiang}]{wizardlm}
Can Xu, Qingfeng Sun, Kai Zheng, Xiubo Geng, Pu~Zhao, Jiazhan Feng, Chongyang Tao, and Daxin Jiang. 2023.
\newblock Wizardlm: Empowering large language models to follow complex instructions.
\newblock \emph{arXiv preprint arXiv:2304.12244}.

\bibitem[{Yadav et~al.(2024)Yadav, Tam, Choshen, Raffel, and Bansal}]{ties}
Prateek Yadav, Derek Tam, Leshem Choshen, Colin~A Raffel, and Mohit Bansal. 2024.
\newblock Ties-merging: Resolving interference when merging models.
\newblock \emph{Advances in Neural Information Processing Systems}, 36.

\bibitem[{Yang et~al.(2024{\natexlab{a}})Yang, Yang, Hui, Zheng, Yu, Zhou, Li, Li, Liu, Huang, Dong, Wei, Lin, Tang, Wang, Yang, Tu, Zhang, Ma, Xu, Zhou, Bai, He, Lin, Dang, Lu, Chen, Yang, Li, Xue, Ni, Zhang, Wang, Peng, Men, Gao, Lin, Wang, Bai, Tan, Zhu, Li, Liu, Ge, Deng, Zhou, Ren, Zhang, Wei, Ren, Fan, Yao, Zhang, Wan, Chu, Liu, Cui, Zhang, and Fan}]{qwen2}
An~Yang, Baosong Yang, Binyuan Hui, Bo~Zheng, Bowen Yu, Chang Zhou, Chengpeng Li, Chengyuan Li, Dayiheng Liu, Fei Huang, Guanting Dong, Haoran Wei, Huan Lin, Jialong Tang, Jialin Wang, Jian Yang, Jianhong Tu, Jianwei Zhang, Jianxin Ma, Jin Xu, Jingren Zhou, Jinze Bai, Jinzheng He, Junyang Lin, Kai Dang, Keming Lu, Keqin Chen, Kexin Yang, Mei Li, Mingfeng Xue, Na~Ni, Pei Zhang, Peng Wang, Ru~Peng, Rui Men, Ruize Gao, Runji Lin, Shijie Wang, Shuai Bai, Sinan Tan, Tianhang Zhu, Tianhao Li, Tianyu Liu, Wenbin Ge, Xiaodong Deng, Xiaohuan Zhou, Xingzhang Ren, Xinyu Zhang, Xipin Wei, Xuancheng Ren, Yang Fan, Yang Yao, Yichang Zhang, Yu~Wan, Yunfei Chu, Yuqiong Liu, Zeyu Cui, Zhenru Zhang, and Zhihao Fan. 2024{\natexlab{a}}.
\newblock Qwen2 technical report.
\newblock \emph{arXiv preprint arXiv:2407.10671}.

\bibitem[{Yang et~al.(2024{\natexlab{b}})Yang, Shen, Guo, Wang, Cao, Zhang, and Tao}]{yang2024model}
Enneng Yang, Li~Shen, Guibing Guo, Xingwei Wang, Xiaochun Cao, Jie Zhang, and Dacheng Tao. 2024{\natexlab{b}}.
\newblock Model merging in llms, mllms, and beyond: Methods, theories, applications and opportunities.
\newblock \emph{arXiv preprint arXiv:2408.07666}.

\bibitem[{Yu et~al.(2024{\natexlab{a}})Yu, Yu, Yu, Huang, and Li}]{yu2024extend}
Le~Yu, Bowen Yu, Haiyang Yu, Fei Huang, and Yongbin Li. 2024{\natexlab{a}}.
\newblock Extend model merging from fine-tuned to pre-trained large language models via weight disentanglement.
\newblock \emph{arXiv preprint arXiv:2408.03092}.

\bibitem[{Yu et~al.(2024{\natexlab{b}})Yu, Yu, Yu, Huang, and Li}]{dare}
Le~Yu, Bowen Yu, Haiyang Yu, Fei Huang, and Yongbin Li. 2024{\natexlab{b}}.
\newblock Language models are super mario: Absorbing abilities from homologous models as a free lunch.
\newblock In \emph{Forty-first International Conference on Machine Learning}.

\bibitem[{Zhang et~al.(2023)Zhang, Liu, He et~al.}]{zhang2023composing}
Jinghan Zhang, Junteng Liu, Junxian He, et~al. 2023.
\newblock Composing parameter-efficient modules with arithmetic operation.
\newblock \emph{Advances in Neural Information Processing Systems}, 36:12589--12610.

\bibitem[{Zhou et~al.(2024)Zhou, Song, Wang, and Chen}]{metagpt}
Yuyan Zhou, Liang Song, Bingning Wang, and Weipeng Chen. 2024.
\newblock Metagpt: Merging large language models using model exclusive task arithmetic.
\newblock \emph{arXiv preprint arXiv:2406.11385}.

\end{thebibliography}

\appendix

\section{Experimental Details}

\subsection{Details of Baselines and Ablation Study}  
\label{appendix:ablation}  

To provide a better understanding of the baselines, we outline the methods used in previous works for addressing intra-model and inter-model interference. We compare these methods with our approach in Table~\ref{tab:comp}, highlighting the innovation of our method. At the bottom of Table~\ref{tab:comp}, we also provide the implementation of the methods used in our ablation study.  

Below is a brief introduction to each component. 
\begin{itemize}  
    \item \textbf{Magnitude Pruning}: Retains parameters with the largest magnitude values.  
    \item \textbf{Random Drop and Rescale}: Filters parameters using a Bernoulli distribution and rescales the remaining ones according to the drop rate.  
    \item \textbf{Stochastic Magnitude Pruning}: Assigns magnitude values to probabilities and retains parameters according to these probabilities.  
    \item \textbf{Disjoint Mean}: Elects parameters at each position based on the direction of summation, then averages the parameters along that direction.  
    \item \textbf{Consensus Mask}: Selects parameters using a mask constructed by measuring the $l_1$ distance to the target task vector.  
\end{itemize}

\begin{table}[h]
\renewcommand{\arraystretch}{1.1}
\resizebox{1.03\columnwidth}{!}{%
\begin{tabular}{c|cc}
\hline
\textbf{Method} & \textbf{Intra-Model}  & \textbf{Inter-Model} \\ \hline
TA            & /            & /               \\
TIES            & Magnitude Pruning            & Disjoint Mean               \\
DARE            & Random Drop and Rescale      & Disjoint Mean                           \\
DELLA           & Stochastic Magnitude Pruning & Disjoint Mean               \\
TALL-Mask       & /                            & Consensus Mask              \\
OBIM            & Saliency-based Pruning       & Iterative Merging           \\ \hline
TIES+OBM        & Saliency-based Pruning       & Disjoint Mean               \\
TIES+IM         & Magnitude Pruning            & Iterative Merging               \\ \hline
\end{tabular}
}
\caption{Comparison of methods for addressing intra-model and inter-model interference.}
\label{tab:comp}
\end{table}

\subsection{Hyperparameter Configurations}  

In the SFT model merging experiments, the hyperparameters that need to be adjusted include the retention ratio of parameters in OBM ($n_k\%$) and the merging order ($O$). The search ranges and the optimal settings for each hyperparameter are provided in Table~\ref{tab:param}.

\begin{table*}[h]
\centering
\resizebox{0.8\textwidth}{!}{%
\begin{tabular}{c|cc}
\hline
\textbf{H-Param} & \textbf{Searching Range}                                                                                                               & \textbf{Optimal Setting}                    \\ \hline
$n_k\%$        & \begin{tabular}[c]{@{}c@{}}{[}0.1, 0.15, 0.2, 0.25, 0.3, \\ 0.35, 0.4, 0.45, 0.5{]}\end{tabular}                              & \{LM: 0.4, Math: 0.45, Code: 0.1\} \\
$O$        & \begin{tabular}[c]{@{}c@{}}{[}Rotation, LM First, Math First, Code First, \\ LM Last, Math last, Code Last{]}\end{tabular}           & Code Last               \\ \hline
\end{tabular}
}
\caption{Hyperparameter search ranges and optimal settings for the SFT model merging experiment.}
\label{tab:param}
\end{table*}

In the post-pretraining model merging scenario, the retention ratio $n_k\%$ for merging $K$ checkpoints is set to \(\frac{1}{K}\), and the merging order is set to \textit{Rotation}.

\subsection{Datasets for Post-Pretraining}  
\label{appendix:pt_data}
We collected and processed a dataset of over 200B tokens comprising Japanese, Chinese, and English texts for post-pretraining. The dataset includes text from websites and publications, all of which are publicly available. Below are the details of the data sources:  

\begin{itemize}  
    \item \textbf{Japanese}: C4-ja\footnote{https://huggingface.co/datasets/systemk/c4-ja}, CC100-ja\footnote{https://huggingface.co/datasets/statmt/cc100}, OSCAR-ja\footnote{https://huggingface.co/datasets/ohtaman/oscar\_ja\_clean\_filtered} CulturaX\footnote{https://huggingface.co/datasets/uonlp/CulturalX}, Wikipedia-ja\footnote{https://huggingface.co/datasets/systemk/wiki-ja}  
    \item \textbf{English}: FineWeb\footnote{https://huggingface.co/datasets/HuggingFaceFW/fineweb}, Tiny-Textbooks\footnote{https://huggingface.co/datasets/nampdn-ai/tiny-textbooks}, AutoMathText\footnote{https://huggingface.co/datasets/math-ai/AutoMathText}, Wikipedia-en\footnote{https://huggingface.co/datasets/blo05/cleaned\_wiki\_en}  
    \item \textbf{Chinese}: CLUECorpus\footnote{https://github.com/brightmart/nlp\_chinese\_corpus}, SkyPile\footnote{https://huggingface.co/datasets/Skywork/SkyPile-150B}, MAP-CC\footnote{https://huggingface.co/datasets/m-a-p/MAP-CC}, Wanjuan\footnote{https://huggingface.co/datasets/facat/wanjuan}, Wikipedia-zh\footnote{https://huggingface.co/datasets/shaowenchen/wiki\_zh}  
    \item \textbf{Parallel Corpus}\footnote{https://opus.nlpl.eu}: CCMatrix, JParaCrawl, WikiMatrix  
\end{itemize}  

We follow the pretraining data processing approach of DataComp-LM~\cite{li2024datacomp}. The data processing pipeline mainly consists of three steps:  

\begin{itemize}  
    \item \textbf{Text Extraction}: We extract clean text from raw content using rule-based tools such as HTML parsers and regular expressions.  
    \item \textbf{Deduplication}: We apply both locality-sensitive hashing (LSH) deduplication and semantic deduplication to remove redundant data.  
    \item \textbf{Quality Filtering}: We employ a \texttt{FastText}\footnote{https://github.com/facebookresearch/fastText} binary classifier for each language to assess and filter data quality.  
\end{itemize}  

After processing, we retain approximately 70B tokens for each language and 1B tokens from the parallel corpus as our post-pretraining dataset.  

We then trained \textit{Qwen2-7B} on this dataset using 64 A800 (80G) GPUs, with a training batch size of 4 million tokens per step for two weeks. Checkpoints were saved every 1,000 steps, and the total training duration was approximately 50,000 steps. 

We will release the trained model as open-source after the paper is accepted.  

\subsection{Details of Validation Set}
\label{appendix:valid_set}

The sample size of the validation set for each benchmark is provided in Table~\ref{tab:validation_set}.

\begin{table}[ht]
\centering
\resizebox{\columnwidth}{!}{%
\begin{tabular}{c|c|c|c}
\hline
\textbf{Model} & \textbf{Benchmark} & \textbf{Sample Size} & \textbf{Total} \\ \hline
\multirow{2}{*}{LM} & AlpacaEval & 50 & \multirow{2}{*}{335} \\
 & MMLU & 285 &  \\ \hline
\multirow{2}{*}{Math} & GSM8K & 50 & \multirow{2}{*}{100} \\
 & MATH & 50 &  \\ \hline
\multirow{2}{*}{Code} & HumanEval & 0 & \multirow{2}{*}{50} \\
 & MBPP & 50 &  \\ \hline
\multirow{3}{*}{\begin{tabular}[c]{@{}c@{}}Post-pretrained \\ Model\end{tabular}} & MMLU & 285 & \multirow{3}{*}{763} \\
 & C-Eval & 260 &  \\
 & JP-LMEH & 218 &  \\ \hline
\end{tabular}
}
\caption{Validation Set Sizes. Note that HumanEval is excluded since it only provides a test set.}
\label{tab:validation_set}
\end{table}

\subsection{Computational Resources}

We measured the GPU usage and time cost in OBIM, as shown in Table~\ref{tab:computational_resources}. GPUs used in our experiments is NVIDIA-A800 (80G). The total time is divided into three parts: the computation time for $\mathbf{X}^l$ across all layers, which depends on the size of the validation set; the time required for saliency computation; and the time for iterative merging.

\section{Results on Merging Multiple Models}
\label{appendix:multi}
Table~\ref{tab:scalability_study} presents the full results for merging varying numbers of post-pretrained models using OBIM and DARE.

\begin{table}[hb]
\setcounter{table}{10}
\centering
\resizebox{\columnwidth}{!}{%
\begin{tabular}{c|c|c|c|c|c}
\hline
\textbf{Count} & \textbf{Method} & \textbf{CEVAL} & \textbf{MMLU} & \textbf{JP-LMEH} & \textbf{Avg.} \\ \hline
\multirow{2}{*}{2} & OBIM & \textbf{81.87} & \textbf{69.41} & 71.85 & \textbf{74.38} \\
 & DARE & 81.20 & 69.16 & \textbf{71.92} & 74.09 \\ \hline
\multirow{2}{*}{3} & OBIM & \textbf{80.98} & 69.32 & \textbf{71.80} & \textbf{74.03} \\
 & DARE & 79.79 & \textbf{69.51} & 71.02 & 73.44 \\ \hline
\multirow{2}{*}{4} & OBIM & \textbf{80.46} & \textbf{69.89} & \textbf{72.14} & \textbf{74.16} \\
 & DARE & 78.97 & 68.85 & 71.21 & 73.01 \\ \hline
\multirow{2}{*}{5} & OBIM & \textbf{79.72} & \textbf{69.40} & \textbf{71.90} & \textbf{73.69} \\
 & DARE & 79.19 & 68.98 & 70.92 & 73.03 \\ \hline
\multirow{2}{*}{6} & OBIM & \textbf{79.72} & \textbf{69.66} & \textbf{71.90} & \textbf{73.76} \\
 & DARE & 78.90 & 69.05 & 70.99 & 72.98 \\ \hline
\end{tabular}%
}
\caption{Detailed performance comparison of OBIM and DARE across different numbers of merged models.}
\label{tab:scalability_study}
\end{table}

\begin{table*}[t]
\setcounter{table}{9}
\centering
\resizebox{\textwidth}{!}{%
\begin{tabular}{c|c|c|c|c|c}
\hline
\textbf{Model Count} & \textbf{Model Size} & \textbf{GPUs} & \textbf{Computing Speed of $\mathbf{X}^l$} & \textbf{Saliency Computation Time} & \textbf{Merging Time}  \\ \hline
3 & 13B & 2 & 0.40 (s/sample) & 50 (s) & 2'03 \\ \hline
6 & 7B & 1 & 0.36 (s/sample) & 20 (s) & 2'01 \\ \hline
\end{tabular}%
}
\caption{Computational resources for model merging with OBIM.}
\label{tab:computational_resources}
\end{table*}

\section{Ethics Statement}

This paper focuses on model merging techniques for Large Language Models (LLMs) to enhance their adaptability. While our work does not directly introduce new risks, it inherits the broader societal concerns associated with LLMs, such as AI safety, reliability, and potential biases in generated content. Beyond these considerations, we do not foresee additional ethical concerns arising from our work.

\end{document}